\documentclass[runningheads]{llncs}
\usepackage{graphicx}
\usepackage{amsfonts}
\usepackage{amsmath}
  \usepackage[
    backend=biber,
    style=nature,
  ]{biblatex}
\usepackage[usenames,dvipsnames]{xcolor} % REMOVE FOR SUBMISSION

\usepackage{biblatex}
\bibliography{biblio}

\newcommand{\unaryminus}{\scalebox{0.30}[1.0]{\( - \)}}

\begin{document}
\title{
Can I trust my anomaly detection system?\\
A case study based on explainable AI.
%\thanks{Supported by the EU project DistriMuSe (KDT). 
% \red{check if we want to do that...}}
% \\ \small{(submitted for review as short paper)}
}
\titlerunning{Can I trust my anomaly detection system? A case study based on XAI.}

\author{
Muhammad Rashid\inst{1}\orcidID{0000{\unaryminus}0002{\unaryminus}2557{\unaryminus}6845}, Elvio Amparore\inst{1}\orcidID{0000{\unaryminus}0003{\unaryminus}1147{\unaryminus}8985},
\\
Enrico Ferrari\inst{2}\orcidID{0000{\unaryminus}0002{\unaryminus}0666{\unaryminus}6597},
Damiano Verda\inst{2}\orcidID{0000{\unaryminus}0001{\unaryminus}9912{\unaryminus}3454}
}

\authorrunning{M. Rashid, and E.G. Amparore, and E. Ferrari and D. Verda}
\institute{
   University of Torino,~ Computer Science Department,\\
   C.so Svizzera 185,~ 10149 Torino,~ Italy\\
   \email{\{muhammad.rashid, elviogilberto.amparore\}@unito.it}
   \vspace{4pt}
\and
    Rulex Innovation Labs, 
    Via Felice Romani 9,~ 16122 Genova,~ Italy\\
    \email{\{enrico.ferrari, damiano.verda\}@rulex.ai}
}

\maketitle              % typeset the header of the contribution
\begin{abstract}

Generative models based on variational autoencoders are a popular technique for detecting anomalies in images in a semi-supervised context.
A common approach employs the anomaly score to detect the presence of anomalies, and it is known to reach high level of accuracy on benchmark datasets.
However, since anomaly scores are computed from reconstruction disparities, they often obscure the detection of various spurious features, raising concerns regarding their actual efficacy.

This case study explores the robustness of an anomaly detection system based on variational autoencoder generative models through the use of eXplainable AI methods.
The goal is to get a different perspective on the real performances of anomaly detectors that use reconstruction differences. 
In our case study we discovered that, in many cases, samples are detected as anomalous for the wrong or misleading factors.

\keywords{anomaly detection  
\and variational autoencoder 
\and eXplainable AI.}
\end{abstract}

\section{Introduction}

The popularity of machine learning methods in difficult tasks, like the  detection of anomalies in industrial quality-control processes, has witnessed a significant surge over the past decade.
Variational AutoEncoders paired with a Generative Adversarial Networks, commonly referred as VAE-GAN\,\cite{vae-gan:larsen16} models, are particularly prominent in this regard, due to their high potential in representation learning.
Anomaly Detection~(AD) on image data with Deep Generative Models~(DGM)\,\cite{zhou2017anomaly} operates on the premise that a model can be trained to learn a representation of the normal features of a sample, while deliberately excluding the capacity to represent and generate any anomalies. 
An \emph{anomaly score} can then be defined on the difference between the original image and its reconstruction, thus quantifying the representational gap for the sample abnormalities.

While successful results have been reported using this strategy\,\cite{ravi2021general}, significant challenges remain. 
An important issue with this approach is that reconstruction differences may actually be either real anomalies, or could be caused by the inability of the generative model to faithfully reproduce the input image.
Additionally, VAE-GAN models often produce images that lack sharpness and details, amplifying differences, particularly at the borders.
Even VAE model with vector quantization exhibit limited improvement in the reconstruction task\,\cite{vqvae2017neural}.

This paper presents a small case study of the performances of a VAE-GAN AD system applied on the popular MVTec dataset\,\cite{bergmann2021mvtec}. 
We review the general framework for anomaly detection using autoencoders by Ravi\,\&\,al.\,\cite{ravi2021general}, which was outlined qualitatively but lacked quantitative evaluation.  
Our study reproduces that framework, augmenting it with additional insights for the explanation part. 
The work of~\cite{ravi2021general} leveraged eXplainable AI (XAI) techniques like LIME and SHAP specifically adapted for anomaly detection (AD). However, their focus was on using XAI for visual explanation to improve anomaly localization compared to basic residual maps, rather than ensuring that the explained anomalies themselves were valid. Additionally, they did not quantify their findings.

In this paper we:
\begin{itemize}
    \item Review an explainable AD system architecture that combines VAE-GAN models with the LIME and SHAP explanation methods;
    \item Quantify the AD system efficacy using anomaly scores;
    \item Use XAI methods to determine if anomalies are indeed detected for the right reason by comparing them with a ground truth, improving the framework of ~\cite{ravi2021general}.
    Our results reveal instances where samples were classified as anomalous but for incorrect reasons. To identify such samples, we employ a methodology based on the optimal Jaccard score. 
\end{itemize}

\section{Literature review}

AD is a well developed field, that has received a lot of attention due to its critical role in numerous practical applications.
Creating effective detection systems is challenging due to several factors, like the difficulty of precisely define what an abnormality is within specific contexts, or the the lack of anomalous samples.

For these reasons, explaining the behaviour of an AD system remains a complex task.
While general purpose interpretability techniques such as GradCAM\,\cite{selvaraju2017grad}, LIME\,\cite{ribeiro2016should} or SHAP\,\cite{lundberg2017unifiedSHAP,rozemberczki2022shapley}  are available, some scholars regard them as imprecise and unreliable\,\cite{kascenas2023anomaly}.
Moreover, their application in the realm of anomaly detection is inherently challenging, due to the lack of a probabilistic black-box function to explain.
Nonetheless, these methodologies can be adapted to offer invaluable insights into understanding the rationale behind the behavior of AD systems. 
In this study we focus on LIME and SHAP systems, due to their (partially) comparable characteristics and their capability in localizing activation areas in anomaly maps.
A broader recent review on AD systems is \cite{Liu2024DeepADSurvey}.

An XAI method for VAE-based systems is VAE-LIME\,\cite{schockaert2020vaelime}, which is based on generating random samples in the latent space of the VAE model. However, it is unclear how this approach can be used in an anomaly detection setting, as it is not obvious how perturbed latent dimensions maps back to the original image segments.
A methodology for explaining anomalies detected by VAE models using SHAP has been developed in~\cite{shapExplainingAnomalies2019}, and our study considers this approach.

A general anomaly detection framework using autoencoders for images is discussed qualitatively in~\cite{ravi2021general}, with our study focusing on reproducing and refining it, particularly in the explanation aspect. 
In that framework, AD relies on anomaly scores, requiring threshold calibration. 
The challenges of perturbation-based methods, such as difficulty in setting appropriate thresholds, are addressed in~\cite{tritscher2023feature}. 
Alternatives like residual explainers for AD have been explored in~\cite{oliveira2021new}.

In XAI for anomaly detection, \emph{anomaly localization}\,\cite{venkataramanan2020attention} is crucial. It improves interpretability by transitioning from pixel-based scores to region localization, especially challenging given the small size of anomalies in real-world datasets.

\begin{figure}[t]
    \centering
    \includegraphics[width=\columnwidth]{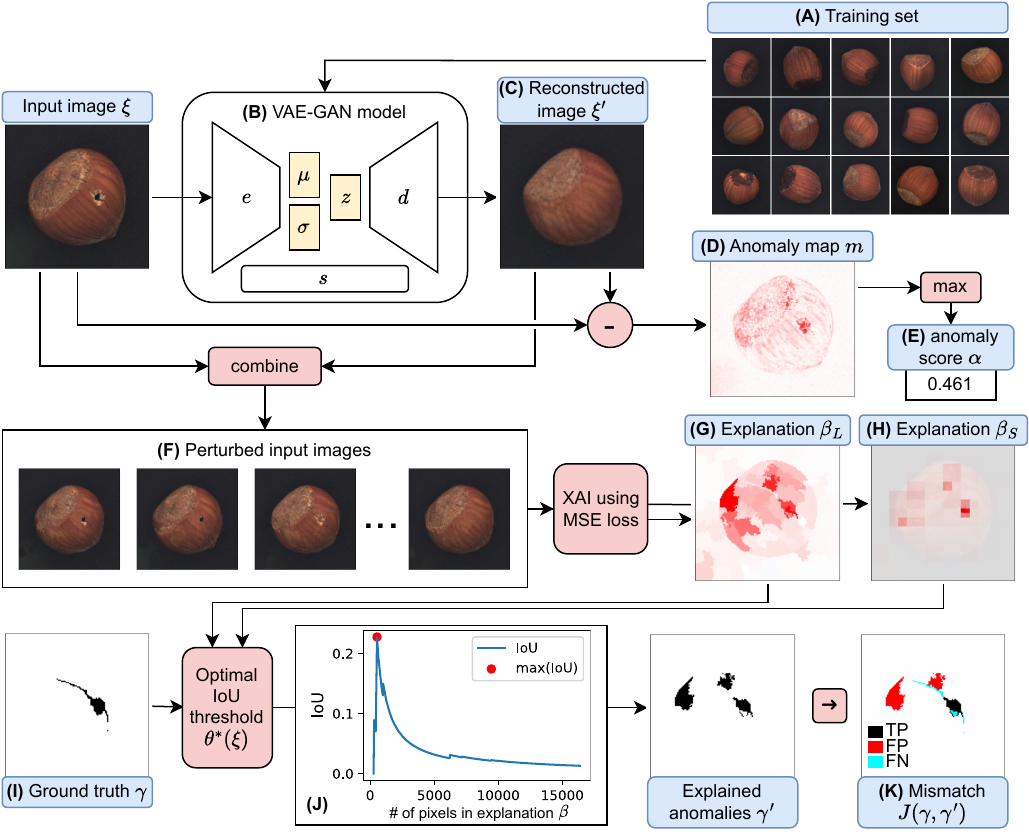}
    \caption{AD system using a VAE-GAN model with LIME explanations.}
    \label{fig:workflow}
\end{figure}

\section{Preliminaries}

We describe the relevant preliminaries following the workflow depicted in Fig.~\ref{fig:workflow}. 
The approach shares many similarity with~\cite{ravi2021general}.
Consider the problem for the domain of $h \times w$ images $\mathcal{I} \in [0-255]^{h \times w \times 3}$, where a sample $\xi \in \mathcal{I}$ may be normal or anomalous. 
We consider images from the high-quality open industrial dataset MVTec\,\cite{bergmann2021mvtec}, namely the categories \emph{hazelnut} and \emph{screw}. 
From a training set (Fig.\,\ref{fig:workflow}/A) containing only normal data (i.e. without anomalies) a VAE-GAN model is trained (Fig.\,\ref{fig:workflow}/B).

\subsection{VAE-GAN models}

A Variational Autoencoder Generative Adversarial Network (VAE-GAN) combines\,\cite{vae:kingma2014auto,vae-gan:larsen16} the strengths of both variational autoencoders (VAEs) and generative adversarial networks (GANs)\,\cite{goodfellow2014generative}.
A VAE-GAN consists of an encoder $e$, a decoder $d$ and a discriminator $s$.
The encoder function $e : \mathcal{I} \rightarrow \mathcal{Z}$ maps input data, such as images $\mathcal{I}$, to a lower-dimensional latent space $\mathcal{Z} \in \mathbb{R}^z$, where each point in $\mathcal{Z}$ represents a potential data sample.
The decoder function $d : \mathcal{Z} \rightarrow \mathcal{I}$ estimates a potential input from a latent space representation, i.e. $d$ approximates $e^{-1}$.
Therefore, encoding and decoding an input image $\xi$ results in its reconstruction (Fig.\,\ref{fig:workflow}/C) through the latent representation $z$, given by
$$
    z = e(\xi), \qquad  \xi' = d(z)
$$
The distribution of the latent space is learnt using a probabilistic approach, and adopts both a regularization of the latent distribution (usually Gaussian) and a GAN approach for adversarial (joint) training of both $d$ and $e$ using the discriminator function $s$ (a classifier trained to distinguish between real and generated data).
When encoded, each data point is described by a Gaussian distribution, with mean $\mu$ and (log)-variance $\sigma$, from which new samples $z$ can be drawn.

\subsection{Semi-supervised anomaly detection using variational models}

While the task of identifying anomalies, particularly in image-based data, holds significant interest across various application domains\,\cite{bergmann2021mvtec,CHOW2020101105}, creating effective anomaly detectors remains a challenge.
Imbalanced datasets are common, with anomalous data being significantly underrepresented (due to the infrequency of anomalous events).
Furthermore, the definition of what constitutes an anomaly is often ambiguous, making supervised learning approaches impractical. 
% Moreover, it is usually unclear what an anomaly should be, making supervised learning not viable.
Therefore, a relevant approach is based on the use of \emph{semi-supervised} learning, where models are trained to detect anomalies from ``normal'' data only. 
Several approaches are possible to perform anomaly detection in a semi-supervised way\,\cite{anomalyDetectionReview}, and in this study we consider a VAE-GAN-based approach\,\cite{an2015variational}.

A VAE-GAN model $(e,d,s)$ for AD is trained exclusively on ``normal'' data, ensuring that only normal data has a proper representation in the latent space.
Consider an input image $\xi$, and let $\xi' = d(e(\xi))$ be its encoding-decoding through the VAE-GAN model. 
If the sample is normal and lies in-distribution with the model, it should be reconstructed accurately, with minimal reconstruction errors.
Conversely, if $\xi$ has anomalous regions, its reconstruction $\xi'$ is likely to resemble that of a normal sample, thereby allowing anomalies to be detected by difference.

Following \cite{ravi2021general}, an \emph{anomaly reconstruction error map} $m \in \mathbb{R}^{h \times w}$ assigns to each pixel of an image $\xi$ its likelihood of being anomalous (Fig.\,\ref{fig:workflow}/D), using 
$$
    m = \bigl|  gs(\xi) - gs(\xi'  \bigr|,
    \qquad\qquad
    \alpha = \max(m)
$$
where $gs : \mathbb{R}^{h \times w \times 3} \rightarrow \mathbb{R}^{h \times w}$ performs a per-pixel maximization of the three color channel values, $\alpha$ is the maximum anomaly value found, denoted as the \emph{anomaly score} (Fig.\,\ref{fig:workflow}/E).
Alternative definitions of anomaly scores have also been explored\,\cite{shapExplainingAnomalies2019}.
While the anomaly map $m$ can be used to visually inspect the reconstruction error, it suffers from limitations:
\begin{itemize}
    \item it does not distinctly identify the anomaly per se, being at the pixel level;
    \item it provides only superficial insights into why a sample may be deemed anomalous.
\end{itemize}

An \emph{anomaly detection threshold} $\tau$ is used to decide if a sample is classified as anomalous, i.e. when $\alpha \geq \tau$.
An \emph{optimal threshold} $\tau^*$ for the whole dataset can be determined using a calibration set (in this study, the test set) as
$$
    \tau^* = \underset{\tau}{\text{argmax}}\,
        \sqrt{ \text{TPR}(\tau)
        \times (1 - \text{FPR}(\tau))}
$$
where TPR and FPR denote the true positive rate and false positive rate, respectively, for the anomaly detection on the calibration set.
Note that this threshold calibration is a critical and fragile part of this class of AD systems, as it is hard to generalize across different domains or datasets.

\subsection{Explaining anomaly maps using model-agnostic XAI methods}

While anomaly maps reveal the reconstruction errors, they only provide a superficial indication of potential anomaly areas within the input image, lacking precise localization of anomalies. 
To address this limitation, XAI methods have been adopted to help in localizing these areas for anomalous samples. 
We focus on model-agnostic methods based on perturbations of input data.
Although many XAI methods rely on classifier predictions,  reconstruction-based AD does not inherently provide such probability scores, and a special setup is needed\,\cite[4.1]{ravi2021general}. 
We consider two XAI methods, LIME and SHAP, adapted as described.

\subsubsection{LIME.}

The Local Interpretable Model-Agnostic Explanations\,\cite{ribeiro2016should}, is a method for explainable AI that works by creating a simpler, interpretable model that approximates the behavior of a more complex model in a synthetic neighborhood of a particular instance being explained.
Let $f : \mathcal{I} \rightarrow \mathbb{R}$ be a prediction regression function that assigns probability scores to input images $\xi \in \mathcal{I}$.
LIME produces an \textit{high-level explanation} consisting of feature attributions (i.e. real-valued scores) assigned not at the pixel-level, but at the level of $k \ll (w \cdot h)$ \textit{superpixels}.
These superpixels represent pre-determined regions of the input image $\xi$ characterized by a combination of color and spatial continuity. 
A common algorithm used to identify superpixels is \emph{Quickshift}~
\cite{vedaldi2008quickshift}.

The $k$ superpixels are used for masking, which is the step that generates the synthetic neighborhood $\mathcal{N}(\xi)$ made of perturbed images (Fig.\,\ref{fig:workflow}/F). 
A mask $x \in \{0,1\}^k$ is a binary vector representing whether each of the $k$ superpixels should be kept (value 1) or replaced (value 0).
In standard LIME, masking vectors are sampled from an unbiased Bernoulli distribution $B$ having probability $0.5$, but more advanced sampling strategies have been proposed\,\cite{stratifiedLIME2024}.

Let $\xi_x$ be the perturbation of image $\xi$ according to the masking vector $x$. 
The synthetic neighborhood $\mathcal{N}(\xi) = \bigl\{ \xi_x ~|~ x \in X \bigr\}$ is then generated from a set $X$ of $n$ masking vectors, resulting in the corresponding dependent variables $Y = \bigl\{ f(\xi_x) ~|~ \xi_x \in \mathcal{N}(\xi) \bigr\}$.

As previously mentioned, LIME is designed to explain a prediction function $f$, and it is not directly applicable to AD, since there is no function $f$ producing probability scores. 
Nonetheless, it can be used to explain the reconstruction error as follows.
A perturbed image $\xi_x$ for mask $x$ is defined, for every pixel $p$, as
$$
    \xi_x[p] = \begin{cases}
        \xi'[p] & \text{if pixel $p$ belongs to a masked superpixel in $x$} \\
        \xi[p] & \text{otherwise}
    \end{cases}
$$
where the reconstruction error of $\xi_x$ is measured as the mean squared error w.r.t. the original input $\xi$, as
$$
    f(\xi_x) = \text{MSE}(\xi - \xi_x)
$$

An explanation in LIME (Fig.\,\ref{fig:workflow}/G) is obtained by fitting a simple linear model: $Y = X \cdot b + \epsilon$, where the vector $b$ represents the weighted least squares estimator of the regression coefficients of $Y$ on $X$, weighted by an appropriate distance function.
A linear function $g(x)$  with coefficients $b$ acts as a local approximation of the square loss function $f$, and the real coefficients $b[i]$ for each superpixel $1 \leq i \leq k$ are interpreted as \emph{feature attribution} scores.
An image-level \emph{feature attribution explanation} $\beta_L$ assigns feature attribution scores to individual pixel, such that each pixel of the $k$ superpixels receive the corresponding coefficient in $b$.

\subsubsection{SHAP.}

The SHapley Additive exPlanation method\,\cite{lundberg2017unifiedSHAP,fumagalli2024shap} provides a game-theore\-tical approach to assign feature importance scores to an input classified by a black-box model.
Similarly to LIME, it is based on the concept of generating perturbations of the original input (with features masked using one or more ``background'' values). 
In the \emph{KernelSHAP} method, perturbations are drawn from the Shapley distribution function. However, unlike LIME, explanation scores are computed from the marginal contribution that each input feature brings to the explained function $f$.
The \emph{SHAP partition explainer}\,\cite{lundberg2017unifiedSHAP} is a specialized image method that employs a recursive cut approach to localize relevant features within an input image. 
An explanation $\beta_S$ generated by the SHAP partition explainer assigns feature attribution scores directly to pixels (Fig.\,\ref{fig:workflow}/H). 
The granularity of these scores depend on a budget of $n$ perturbed images that the XAI method can produce to explain an input sample $\xi$.

The application of SHAP to explain the anomalies revealed by an autoencoder has been developed in \cite{shapExplainingAnomalies2019} and, similarly to LIME, is based on a reconstruction error function $f(\xi)$ but without relying on any predetermined superpixels.

\subsection{Comparing explained anomalies against a ground truth}

A pixel-level feature attribution explanation $\beta$ generated by an XAI method is a real matrix of feature attribution scores assigned to the pixels of the image.
To assess the method's capability of localizing the anomalous regions in an input image, we adopt the following methodology.
A Boolean ground truth $\gamma \in \{0,1\}^{h \times w}$ is a matrix that assigns, to each pixel of the input image $\xi$, a value whether the pixel belongs to the anomaly being localized or not (Fig.\,\ref{fig:workflow}/I).

We assume that $\gamma$ is available for the anomalous samples of the test set.
Since the explanation $\beta$ is a real-valued matrix, it is not directly comparable with $\gamma$.
An effective way to perform such comparison is to define an \emph{explanation threshold} $\theta$, and define a boolean explanation $\gamma'$, derived from $\beta$, that marks as anomalous those pixels of $\xi$ for which the feature attribution score in $\beta$ is greater than $\theta$.
A comparison between $\gamma$ and $\gamma'$ can then be performed using standard metrics like the Jaccard coefficient (a.k.a. Intersection over Union - IoU)
$$
    J(\gamma, \gamma') = \frac{\gamma \wedge \gamma'}{\gamma \vee \gamma'}
$$
However, determining an optimal threshold $\theta$ is not straightforward. 
Hence, we select, for each explained sample, a corresponding optimal threshold $\theta^*(\xi)$ for which $J(\gamma, \gamma')$ is maximal (Fig.\,\ref{fig:workflow}/J). 
The mismatch between $\gamma$ and $\gamma'$ can then be inspected and visualized\footnote{We adopt a threshold-maximization approach instead of a threshold-independent metric like AU-IoU, because the former has a more intuitive visualization.} (Fig.\,\ref{fig:workflow}/K).
Note that this coefficient can only be computed when $\gamma$ is available and it is not empty (otherwise it would be meaningless). Thus it can be used only to explain anomalies for ``abnormal'' samples, but it cannot be used on ``good'' samples.

\begin{figure}[t]
  \centering
  \includegraphics[width=\textwidth]{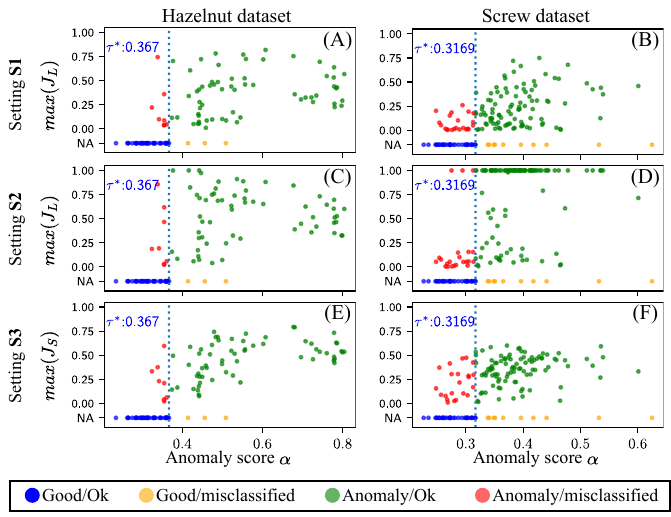}
   \caption{Maximum IoU vs the anomaly scores in the two test datasets.}
   \label{fig:resultsIoUAS}
\end{figure}

\section{Experimental evaluation}

We present the results on a set of experiments made on the MVtec dataset\,\cite{bergmann2021mvtec} and considering two categories, \emph{hazelnut} and \emph{screw}, each comprising images of these objects with and without defects.
The tests use a VAE-GAN model implemented in Keras\,\cite{vaegan:code}, where the encoder model $e$ uses 4 nested convolutional layers ($3{\times}3$ kernel, stride 2), with each layer using ReLU activation and followed by a batch normalization, and using a final Dense decision layer. 
The discriminator $s$ is similar to $e$, but using three convolutional layers with larger kernels ($8{\times}8$, $5{\times}5$ and $4{\times}4$, respectively) and followed also by max pooling.
The decoder $d$ mirrors the structure of $e$, but in reverse order and using transposed convolutions.
Input images are scaled to $128 \times 128$. 
Training is performed on 30\,000 epochs on batches of 64 images, incorporating mild augmentation techniques (rotation, width/height shift, brightness adjustment, zoom) to mitigate overfitting and make the model more robust to variations in background light and shadows.

Due to the dependency of LIME on the quality of the segmentation in superpixels, we consider three  evaluation setups:
\begin{itemize}
    \item \textbf{S1}: LIME explanations with segmentation performed on the input image,
            without prior knowledge of the anomalies (fair setup).
            Potential misbehaviors may arise from either LIME's failure to localize anomalies or inaccuracies in the segmentation method in identifying anomaly boundaries.
            All explanations are computed using $k{=}100$ segments, $n{=}5\,000$ samples. 
    \item \textbf{S2}: LIME explanations with segmentation performed knowing both the image and the ground truth. 
            In this setup, we remove the segmentation method as a potential cause of LIME misbehaviors (anomalies fall into distinct segments).
             However, this setup is unrealistic since it exposes the ground truth.
            As before, we use $k{=}100$ segments and $n{=}5\,000$ samples.
            
    \item \textbf{S3}: SHAP explanations using partition explainer, with $n{=}5\,000$ samples.
\end{itemize}

\begin{figure}[t]
\includegraphics[width=1\linewidth]
{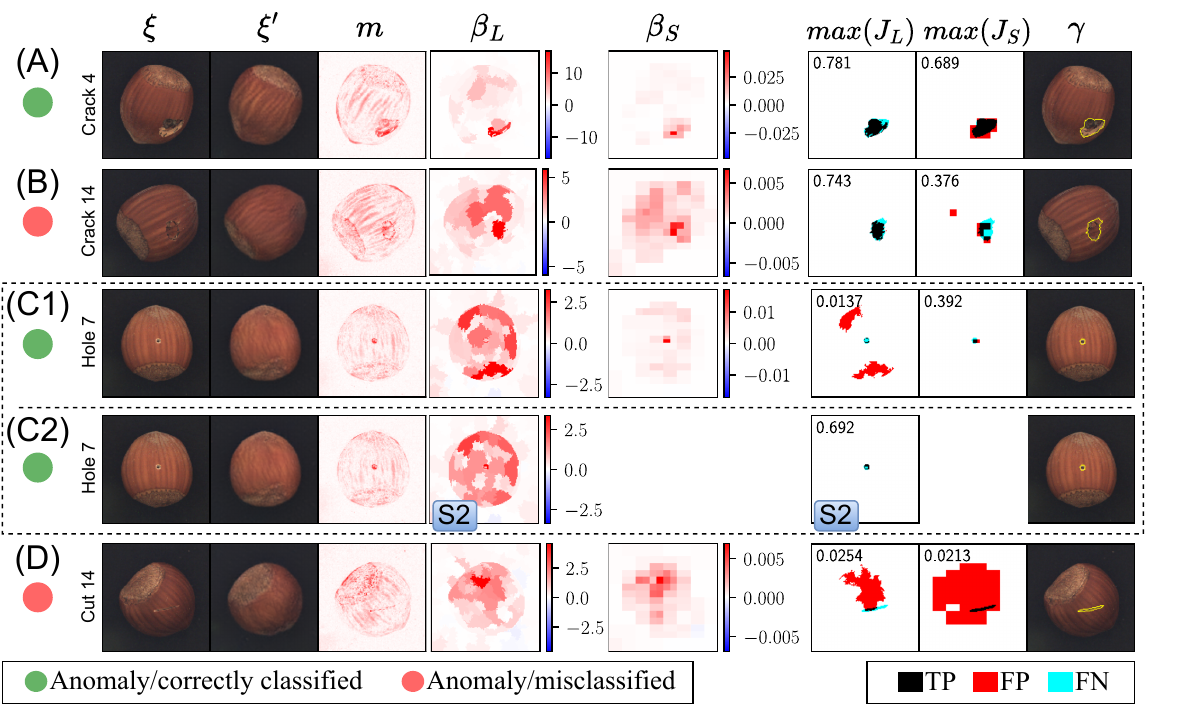}
\caption{Explanations for a few anomalous samples of the hazelnut dataset.} 
\label{fig:hazelnut_samples}
\end{figure}

Explaining using $n{=}5\,000$ samples takes about $20$ seconds on a M1 laptop.
The plots in Fig.~\ref{fig:resultsIoUAS} illustrate the performance of the AD system (X axis) and its explainability in terms of maximal $J(\gamma, \gamma')$ scores (Y axis) on the test sets of the two considered datasets (left and right columns) in the three setups (rows).
We denote LIME and SHAP explanations with $J_L$ and $J_S$, respectively.
Anomaly scores remain consistent within each column, with only the maximal $J_L$ (resp. $J_S$) scores varying.
The hazelnut dataset comprises 40 good (37 correctly classified, 3 misclassified) and 70 anomalous (62 correctly classified, 8 misclassified) samples, reaching 90\% accuracy using the optimal threshold.
The screw dataset includes 41 good (31 correctly classified, 10 misclassified) and 119 anomalous (97 correctly classified, 22 misclassified) samples, achieving 80\% accuracy using the optimal threshold.

While it is expected that the maximal IoU should not be perfect, the scores obtained from the XAI methods already reveal that some samples exhibit very poor localization of the anomalies.
Given that both LIME and SHAP compute explanations based on residual reconstruction errors, it is plausible that some samples are classified as good or anomalous for incorrect reasons. To evaluate this, we conduct manual inspection of the samples.
\medskip

\noindent\textit{Hazelnut dataset.}
Fig.~\ref{fig:hazelnut_samples} illustrates a few selected anomalous samples from the hazelnut dataset\footnote{All test sample explanations are provided separately (link at the end of the paper).}. 
Each row shows, from left to right, the sample $\xi$ and its reconstruction $\xi'$,  the anomaly reconstruction error map $m$, the explanations $\beta_L$ and $\beta_S$ generated from LIME and SHAP, resp., the visualization of the maximal $J_L$ and $J_S$ for both explanation methods (the $J$ value is reported in the upper-left corner), and the boundary of the ground truth region $\gamma$.
All LIME explanations $\beta_L$ come from the S1 setup, unless explicitly labeled as S2. SHAP explanations $\beta_S$ are computed using the S3 setup.
\\
Sample (A) from Fig.~\ref{fig:hazelnut_samples} shows a case of a hazelnut with a small surface crack that is properly localized and detected (with some negligible mistakes).
\\
Sample (B) looks similar, but it is misclassified as good, having the anomaly score $\alpha$ below the threshold $\tau^*$. However, the XAI methods would still localize the anomalous region.
\\
In (C1), $\beta_L$ shows significant confusion, attributing large values to the border instead of the small hole at the center. The primary issue lies in the segmentation: employing a segmentation that accurately encloses the anomaly (as in C2 with the S2 setup) results in better localization(even if some confusion remains). 
This underscores how LIME can be greatly influenced by inadequate segmentation.
% Sample (C1) shows an example where LIME explanation is very confused, assigning large feature attributions values to the border rather than the small hole at the center. In this case, the primary issue lies in the segmentation: utilizing a segmentation that adequately envelopes the anomaly (run (C2) using the S2 setup) leads to improved localization (even if some confusion remains). 
% This shows that LIME can be severely affected by an inadequate segmentation.
\\
Sample (D) shows an example where both LIME and SHAP fail to identify the anomaly accurately: since the reconstruction $\xi'$ is not entirely faithful, both XAI methods mislocate the anomalous region to the top of the image, overlooking the actual one (a cut on the hazelnut shell).

\begin{figure}[t]
\includegraphics[width=1\linewidth]
{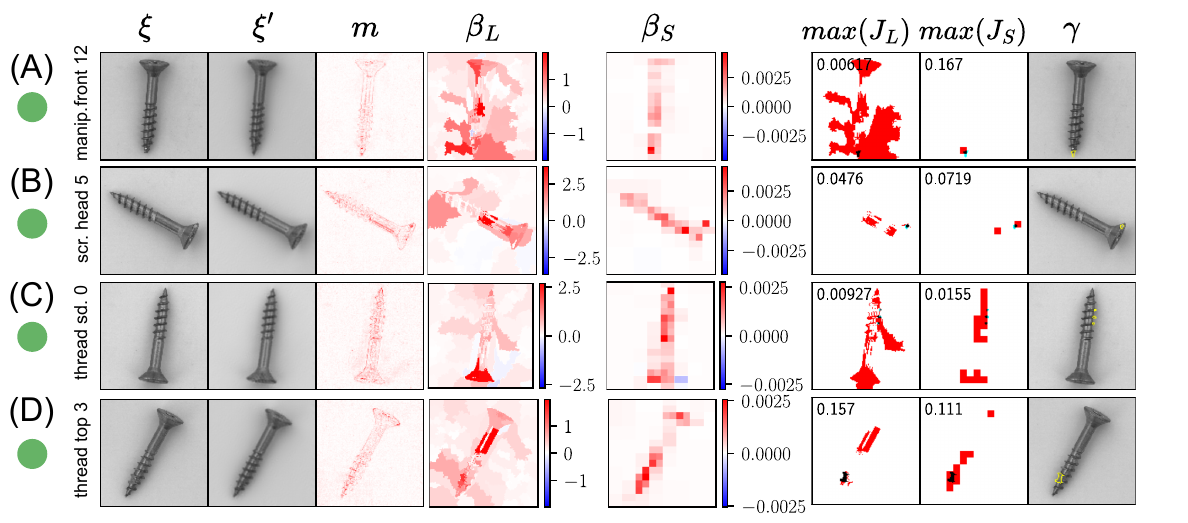}
\caption{Explanations for a few anomalous samples of the screw dataset.} 
\label{fig:screw_samples}
\end{figure}
\medskip

\noindent\textit{Screw dataset.} 
Detecting anomalies in this dataset presents greater difficulty as they typically occupy small portions of the image. While many samples are correctly classified and explained, accurately localizing the anomalous area proves challenging for others.
In the four samples in Fig.~\ref{fig:screw_samples}, all correctly classified as anomalous, the feature attribution scores are maximal in areas that do not contain any anomaly (as evidenced by the large false-positive areas in the $\max(J)$ plots).
Sample (C) is particularly critical, as both LIME and SHAP assign low scores to the region containing the anomaly (the right thread side). 
This suggests that the sample may have been classified as anomalous for the wrong reason, and this could only be detected through the use of XAI methods.

\section{Conclusions}

In this case study we replicated the framework of \cite{ravi2021general}, enhancing it by quantifying both AD and XAI performances. 
Our aim was to highlight the relevance of XAI methods in finding the true drivers behind anomaly detection, particularly when utilizing reconstruction error maps generated from VAE-GAN models.

The results show that relying solely on the anomaly score is insufficient for comprehending the classification process. A sample may be detected as anomalous for the wrong reasons, yet this misbehaviour may not be detectable from the information provided by the anomaly map alone.
We used two model-agnostic XAI methods to obtain explanations from the anomalous samples, to inspect if the anomalies were correctly localized.
Region localization through a XAI method with Jaccard score maximization allows the user to inspect the AD system, identifying potential misbehaviors in the detection and providing a better understanding of the system.

Both tested XAI methods successfully localizes activation regions, with some discrepancies. 
Specifically, LIME exhibited a slightly inferior performance compared to SHAP, attributable to its reliance on a pre-determined segmentation that is not aware of the ML process and does not get any feedback from it. 
This fragility can be seen by the variations between the S1 and S2 test setups (like in Fig.~\ref{fig:hazelnut_samples}/C1-C2).

\medskip
\noindent \textit{Code availability:}
All code needed to replicate the experiments, including all the explanations for all test samples, are available at:\\ \small{\url{https://github.com/rashidrao-pk/anomaly_detection_trust_case_study}}

\begin{credits}
\subsubsection{\discintname}
The authors have no competing interests to declare that are relevant to the content of this article.

\end{credits}

\printbibliography

\end{document}